\pdfoutput=1
\documentclass[11pt]{article}
\usepackage[preprint]{acl}
\usepackage{times}
\usepackage{latexsym}
\usepackage[T1]{fontenc}
\usepackage[utf8]{inputenc}
\usepackage{microtype}
\usepackage{graphicx}
\usepackage{booktabs}
\usepackage{multirow}
\usepackage{amsmath,amssymb}
\usepackage{xcolor}
\usepackage{pifont}
\usepackage{subcaption}
\usepackage{caption}

\newcommand{\oursname}{\textsc{Mezzo}}

\definecolor{oursred}{HTML}{C53030}
\newcommand{\icon}[1]{\raisebox{-0.18ex}{\includegraphics[height=1.55ex]{figs/icon_#1.pdf}}}

\newcommand{\nampCostlydistressLlamaGeneric}{0.27}

\newcommand{\nampSafedistressLlamaGeneric}{0.61}

\newcommand{\nbetweenDroptoxicityLlamaGeneric}{95}

\newcommand{\nbootPairedDiffMax}{2}
\newcommand{\nbootSeMaxLarge}{4.0}
\newcommand{\nbootSeMaxSmall}{7.7}

\newcommand{\nceilLosspolitenessLlamaGeneric}{15}

\newcommand{\nceilLosstoxicityLlamaGeneric}{36}

\newcommand{\neditBTBt}{21}
\newcommand{\neditLlamaGeneric}{36}
\newcommand{\neditLlamaGrammar}{6}
\newcommand{\neditLlamaPreserve}{37}
\newcommand{\neditLlamaProfessional}{63}
\newcommand{\neditMax}{63}
\newcommand{\neditMin}{2}
\newcommand{\neditQwenLGeneric}{10}
\newcommand{\neditQwenLGrammar}{3}
\newcommand{\neditQwenLPreserve}{2}

\newcommand{\nmaxSoftenQwenLGeneric}{0.19}

\newcommand{\nndcgHalftoxicityLlamaGeneric}{0.85}

\newcommand{\nndcgOnepolitenessLlamaGeneric}{0.78}

\newcommand{\nndcgOnetoxicityLlamaGeneric}{0.68}

\newcommand{\nndcgZeropolitenessLlamaGeneric}{0.83}

\newcommand{\nndcgZerotoxicityLlamaGeneric}{0.89}

\newcommand{\norigThreetoxicity}{0.830}

\newcommand{\npriceRsqpoliteness}{0.88}

\newcommand{\npriceRsqtoxicity}{0.92}

\newcommand{\npriceTenHipoliteness}{9.2}

\newcommand{\npriceTenHitoxicity}{10.8}

\newcommand{\npriceTenLopoliteness}{5.6}

\newcommand{\npriceTenLotoxicity}{7.9}
\newcommand{\npriceTendistress}{2.1}

\newcommand{\npriceTenintensity}{2.5}

\newcommand{\npriceTenpoliteness}{7.8}

\newcommand{\npriceTentoxicity}{9.7}

\newcommand{\nrangetoxicityLlamaGeneric}{39}

\newcommand{\nreaderGapClosedtoxicityBTBt}{78}
\newcommand{\nreaderGapClosedtoxicityLlamaGeneric}{6}

\newcommand{\nreaderInformedDeltaMaxLlamaGeneric}{4}

\newcommand{\nreaderNaivepolitenessLlamaGeneric}{0.24}

\newcommand{\nreaderNaivetoxicityLlamaGeneric}{0.19}

\newcommand{\nreaderOrigpoliteness}{0.39}
\newcommand{\nreaderOrigtoxicity}{0.52}

\newcommand{\nreaderRetintensityLlamaGeneric}{98}

\newcommand{\nrecallPoltoxicityLlamaGeneric}{6}

\newcommand{\nrecallUnpoltoxicityLlamaGeneric}{73}

\newcommand{\nrecaltoxicityLlamaGeneric}{0.533}

\newcommand{\nrecovpolitenessLlamaGeneric}{54}

\newcommand{\nrecovtoxicityLlamaGeneric}{39}

\newcommand{\nrefusesentimentLlamaGeneric}{4.4}

\newcommand{\nrefusetoxicityLlamaGeneric}{7.5}

\newcommand{\nremovalelongationLlamaGeneric}{55}

\newcommand{\nremovalexclamationLlamaGeneric}{71}
\newcommand{\nremovalexclamationQwenLGeneric}{8}

\newcommand{\nremovalfirstpersonLlamaGeneric}{5}

\newcommand{\nremovalintensifierLlamaGeneric}{65}

\newcommand{\nremovalmultipunctLlamaGeneric}{84}

\newcommand{\nremovalprofanityLlamaGeneric}{82}
\newcommand{\nremovalprofanityQwenLGeneric}{26}

\newcommand{\nretInclRefusetoxicityLlamaGeneric}{31}

\newcommand{\nretSocialHiBTBt}{100}
\newcommand{\nretSocialHiLlamaGeneric}{101}

\newcommand{\nretSocialHiQwenLGeneric}{105}

\newcommand{\nretSocialLoBTBt}{89}
\newcommand{\nretSocialLoLlamaGeneric}{42}

\newcommand{\nretSocialLoQwenLGeneric}{94}

\newcommand{\nretdistressLlamaGeneric}{86}

\newcommand{\nretempathyLlamaGeneric}{95}

\newcommand{\nretintensityLlamaGeneric}{93}

\newcommand{\nretironyLlamaGeneric}{101}

\newcommand{\nretpolitenessLlamaGeneric}{67}

\newcommand{\nretpolitenessLlamaProfessional}{42}

\newcommand{\nretsentimentLlamaGeneric}{100}

\newcommand{\nrettopicLlamaGeneric}{98}

\newcommand{\nrettoxicityLlamaGeneric}{42}

\newcommand{\nrettoxicityLlamaPreserve}{64}
\newcommand{\nrettoxicityLlamaProfessional}{28}

\newcommand{\nrhoProfanityTox}{0.46}

\newcommand{\nshifttoxicityLlamaGeneric}{0.345}

\newcommand{\nslopeCostlypolitenessLlamaGeneric}{0.42}

\newcommand{\nslopeCostlytoxicityLlamaGeneric}{0.06}

\newcommand{\nslopeSafepolitenessLlamaGeneric}{0.88}

\newcommand{\nslopeSafetoxicityLlamaGeneric}{0.59}

\newcommand{\nsoftenCostlyempathyLlamaGeneric}{0.32}

\newcommand{\nsoftenCostlypolitenessLlamaGeneric}{0.70}

\newcommand{\nsoftenCostlytoxicityLlamaGeneric}{1.88}

\newcommand{\nsoftenSafeempathyLlamaGeneric}{0.22}

\newcommand{\nsoftenSafepolitenessLlamaGeneric}{0.22}

\newcommand{\nsoftenSafetoxicityLlamaGeneric}{0.16}

\newcommand{\nunreachpoliteness}{46}
\newcommand{\nunreachtoxicity}{80}

\newcommand{\nwithinDroptoxicityLlamaGeneric}{11}

\hypersetup{pdftitle={LLM Polishing as Dynamic Range Compression of Social Signals},pdfauthor={Zhimei Chen, Mu Chen}}

\makeatletter
\AtBeginDocument{\g@addto@macro\@maketitle{%
    \begin{minipage}{\textwidth}
  \centering
  \includegraphics[width=\textwidth]{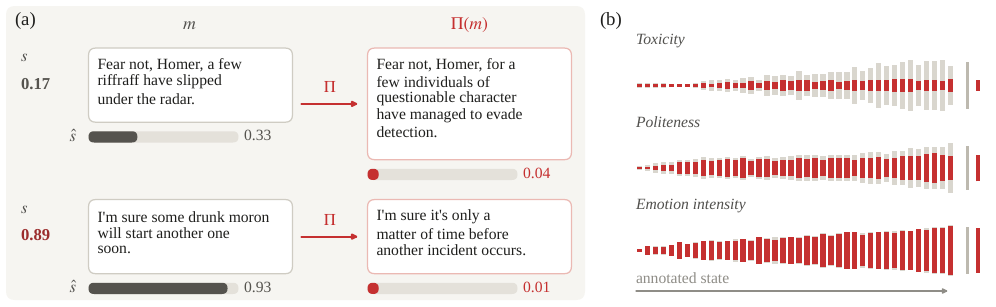}
  \captionof{figure}{Polishing compresses the signals from which readers judge a sender's hostility. (a) A mildly dismissive and a hostile comment read alike once polished. (b) Across all messages, the most toxic and rudest are flattened.}
  \label{fig:teaser}
  \end{minipage}
  \vspace{0.08in}}}
\makeatother

\title{LLM Polishing as Dynamic Range Compression of Social Signals}

\author{%
  Zhimei Chen\textsuperscript{*} \qquad Mu Chen\textsuperscript{1,*,\dag} \\
  \textsuperscript{1}Mohamed bin Zayed University of Artificial Intelligence \\
  \texttt{mu.chen@mbzuai.ac.ae}}

\begin{document}
\maketitle
{\renewcommand{\thefootnote}{\fnsymbol{footnote}}%
\footnotetext[1]{Equal contribution (Zhimei Chen and Mu Chen).}%
\footnotetext[2]{Corresponding author: Mu Chen.}}

\begin{abstract}
Writing assistants polish a growing share of everyday messages before they are read, rewriting them into fluent, courteous prose. Top-leading studies characterize the effect as homogenization, measured as reduced lexical and stylistic diversity of the text that leaves the sender. However, we find that a larger effect lies at the receiver: polishing compresses the signals from which a reader judges how hostile, rude, or distressed the sender is. Each ten percent of a message rewritten costs \npriceTentoxicity\% of the toxicity signal a reader recovers and \npriceTenpoliteness\% of the politeness signal, against under 1\% for irony, empathy, sentiment, and topic. More importantly, even a reader retrained on polished text recovers only \nrecovtoxicityLlamaGeneric\% of the lost toxicity signal, because differently hostile messages become nearly the same text. Across six human-annotated social signals and four open-weight polishers, the loss concentrates on the most hostile and rudest messages and traces to the forms a polisher is asked to remove. Specifically, we introduce \oursname{}, which models polishing as a channel from the sender's state to the reader's inference. When half of the senders polish, a toxicity ranking finds a hostile sender \nrecallUnpoltoxicityLlamaGeneric\% of the time if they wrote unaided and \nrecallPoltoxicityLlamaGeneric\% if they polished.
\end{abstract}

\section{Introduction}

Recently, there has been a trend of routing everyday writing through a large language model (LLM) before it reaches its reader. Writing assistants now sit inside email clients, messaging apps, and phone keyboards; by late 2024 about 18\% of consumer complaints filed with a U.S.\ financial regulator and 24\% of corporate press releases showed signs of LLM assistance \citep{liang2025widespread}, as did about 38\% of posts on question-answering and blogging platforms, against under 3\% on Reddit \citep{sun2025are,lacava2025machines}. A line of work measures what this does to language: writing with an instruction-tuned model reduces the diversity of what a group of writers produces \citep{padmakumar2024does,doshi2024generative}, pulls non-Western writers toward Western style \citep{agarwal2025ai}, narrows the variance of writing complexity and blunts classifiers of author traits \citep{sourati2026shrinking}, and shifts the stance and emotional vocabulary of edited essays in a shared direction \citep{abdulhai2026how}. Polished text, in short, is more uniform than unaided text, and this line of work presumes that what polishing changes can be read off the text that leaves the sender.

However, one concern naturally arises that these measurements describe the message that leaves the sender, while communication succeeds or fails at the receiver. Readers, and the classifiers that platforms run on their behalf, judge from the wording of a message how hostile, rude, or distressed its sender is, and act on that judgment when they moderate, triage, or decide whom to help. Figure~\ref{fig:teaser}(a) tests this presumption directly with two comments annotated 0.17 and 0.89 for toxicity, one mildly dismissive and the other openly hostile. After each writer asks Llama-3.1-8B to ``polish this before I send it,'' ``some drunk moron will start another one soon'' becomes ``it's only a matter of time before another incident occurs'', and a receiver trained on human text, which reads 0.33 and 0.93 from the originals (grey bars), reads 0.04 and 0.01 from the polished versions (red bars). More importantly, even the order of the two writers is lost, since the hostile one is now read as the less toxic. The pair is typical. Figure~\ref{fig:teaser}(b) draws every test message as one bar, ordered by annotated state in 40 equal-count bins and mirrored like an audio waveform, with height equal to the receiver's mean reading of the original (grey) or of the polished message (red), scaled to the range of the original readings: the polished waveform is flattened at the costly end of toxicity and politeness (the latter drawn with rudeness increasing to the right), and follows the original for emotion intensity. A lexical measure would register the more careful vocabulary of both and a diversity measure their similarity, and neither would report the fact that matters to a moderator, who can no longer tell which writer is the hostile one. A polisher is asked to remove what style guides call excessive, such as shouting, profanity, repeated punctuation, and intensifiers, and in human writing these are the conventional carriers of social cost. A drop in textual variance therefore admits two readings with opposite implications: an editor who removes variation that carries no information about the sender makes the sender easier to read, and a polisher that removes the variation that does carry it makes the sender harder to read. Statistics of the text cannot separate the two, and a classifier tested on polished text cannot tell whether what was lost is a distribution shift that a reader could learn to undo or information that is gone.

\begin{figure*}[t]
\centering
\includegraphics[width=\textwidth]{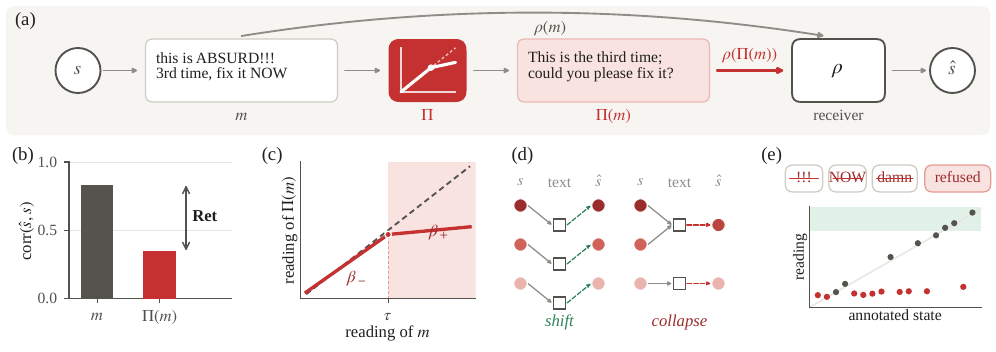}
\caption{\oursname{} compares a receiver's readings of the original and the polished message with the annotated state. (a) Polishing as a channel. (b) Retention. (c) Compression curve. (d) Shift and collapse. (e) Mechanism and ranking.}
\label{fig:framework}
\end{figure*}

In this work, we present \oursname{}~\icon{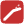}, an evaluation suite that treats polishing as a communication channel and measures what passes through it (Figure~\ref{fig:framework}). We rethink linguistic homogenization in two aspects. The first is \emph{what is measured}: not the marginal statistics of the text but the receiver's inference about the sender's state, summarized, as in audio engineering, by a dynamic-range compressor with a knee and a slope on each side, so that how much is lost becomes where on the scale it is lost. The second is \emph{whether it can be undone}: a loss that a recalibrated reader recovers is a shift~\icon{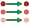}, and a loss that no reader recovers is a collapse~\icon{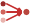}. Concretely, our study has three designs. First, we assemble six social signals with human-annotated ground truth (emotion intensity, politeness, toxicity, irony, empathy, and distress; graded except for irony) and two content controls (sentiment polarity and topic), and pass each message through four open-weight polishers under four requests and through a back-translation control that paraphrases without any model of good writing. Second, we read original and polished messages with trained receivers and a zero-shot LLM reader, so that no conclusion rests on one kind of reader, and quantify retention and the compression curve from original to polished reading. Third, we test invertibility by retraining receivers on polished text and by informing the LLM reader, which separates a shift from a collapse, trace the loss to the surface forms the polisher removes, and simulate a receiver who must rank a pool of senders as the share who polish grows.

Our findings exhibit several compelling facets. First, the loss is selective and priced by rewriting: across fourteen polishing conditions of Qwen2.5-7B, Qwen3-8B, Qwen3-14B, and Llama-3.1-8B, retention falls linearly with the fraction of tokens changed, by \npriceTentoxicity{} points per ten percent for toxicity and \npriceTenpoliteness{} for politeness against under one point for irony, empathy, sentiment, and topic, so that Qwen3-14B, which changes \neditQwenLGeneric\% of the tokens, keeps \nretSocialLoQwenLGeneric{} to \nretSocialHiQwenLGeneric\% of every social signal, Llama-3.1-8B, which changes \neditLlamaGeneric\%, keeps \nrettoxicityLlamaGeneric\% of toxicity, and back-translation, which changes \neditBTBt\%, keeps \nretSocialLoBTBt{} to \nretSocialHiBTBt\%. Second, the loss is concentrated at the costly end: under Llama-3.1-8B the reading of toxicity rises with slope \nslopeSafetoxicityLlamaGeneric{} below the knot and \nslopeCostlytoxicityLlamaGeneric{} above it, the rude quintile of the politeness scale is read \nsoftenCostlypolitenessLlamaGeneric{} sd more polite, and distress is raised at every level, by \nampSafedistressLlamaGeneric{} sd at the calm end, so that the average level of a signal can rise while its range shrinks. Third, the loss is mostly a collapse: a receiver retrained on polished text closes \nrecovtoxicityLlamaGeneric\% of the toxicity gap and \nrecovpolitenessLlamaGeneric\% of the politeness gap, and an informed LLM reader closes \nreaderGapClosedtoxicityLlamaGeneric\%. Fourth, the mechanism is the task itself: Llama-3.1-8B deletes repeated punctuation from \nremovalmultipunctLlamaGeneric\% of the messages that contain it and profanity from \nremovalprofanityLlamaGeneric\%, and declines to rewrite \nrefusetoxicityLlamaGeneric\% of the toxicity messages, almost all at the hostile end, so the most extreme senders are not softened but silenced. Fifth, the consequence scales with adoption and falls on adopters: the NDCG of a toxicity ranking at the top decile falls from \nndcgZerotoxicityLlamaGeneric{} when no sender polishes to \nndcgOnetoxicityLlamaGeneric{} when all do, and when half polish a truly hostile sender is recovered \nrecallUnpoltoxicityLlamaGeneric\% of the time if they did not polish and \nrecallPoltoxicityLlamaGeneric\% if they did. These results suggest that a writing assistant should be evaluated on what it preserves about its user as well as on what it preserves of the content, and we release \oursname{} for this purpose.

\section{Related Work}

\paragraph{AI-Mediated Communication.}
\citet{hancock2020ai} define AI-mediated communication as interpersonal communication in which an intelligent agent modifies, augments, or generates messages on behalf of a communicator. Experiments with reply suggestions find faster and more positive conversations, together with harsher evaluation of a partner suspected of using them \citep{hohenstein2023artificial}; recipients of a message that a friend wrote with AI assistance perceive less effort and report lower relationship satisfaction \citep{liu2023artificial}; and most recipients do not consider that a personal message could be machine-written until told, after which the sender is judged insincere \citep{zhu2026blissful}. These studies, and the loss of signaling value in written applications once generation is cheap \citep{galdin2025making}, concern how a message or sender is perceived once AI use is known or suspected. We hold the sender fixed, assume no disclosure, and ask what a receiver infers about the sender's state from the polished text.

\paragraph{Linguistic Homogenization.}
Writing with an instruction-tuned model reduces the collective diversity of essays and ideas \citep{padmakumar2024does,doshi2024generative,anderson2024homogenization}, language models produce similar outputs across providers \citep{jiang2025artificial}, and suggestions from a Western-centric model pull Indian writers toward Western style \citep{agarwal2025ai}. \citet{sourati2026shrinking} find that the variance of writing complexity fell after the release of ChatGPT, that LLM rewriting cuts it by 21 to 50\%, and that classifiers of author traits trained on original text lose accuracy on rewritten text. \citet{abdulhai2026how} find that edits requested with minimal-change instructions still move essays in a common semantic direction, neutralize their stance, and double the density of emotional vocabulary \citep[see also][]{vannuenen2026voice}. All of these quantify the text. The closest to us is the trait analysis of \citet{sourati2026shrinking}, which measures a receiver-side quantity for stable author traits, with classifiers trained on original text only, so that distribution shift and information loss are not separated. Our dimensions are the communicative states of individual messages, our receivers are retrained to separate the two, and we characterize the shape of the loss.

\paragraph{Machine-Generated Text.}
Population-level estimators and supervised detectors put LLM assistance at 18\% of consumer complaints, 24\% of press releases, a rising share of scientific papers, and about 38\% of posts on Medium and Quora against under 3\% on Reddit \citep{liang2025widespread,liang2025quantifying,sun2025are,lacava2025machines}. These rates set the scale of the collective effects in Section~\ref{sec:collective}; we do not detect polished text, since our question is what is lost even when nobody detects it.

\paragraph{Social Signals in Language.}
Language carries, besides propositional content, graded signals about the speaker, among them emotion intensity, politeness, offensiveness, irony, empathy, and distress, each with an annotated resource that we use as ground truth (Section~\ref{sec:setup}). LLM transformations of clinical text bias uncertainty toward certainty \citep{du2026possible}, a domain-specific instance of the compression we document.

\section{\oursname{}}
\label{sec:method}

This section formalizes polishing as a channel in \S\ref{sec:problem}, defines the measurements of \oursname{} in \S\ref{sec:measures}, and describes the dimensions, polishers, and receivers that instantiate them in \S\ref{sec:setup}. Figure~\ref{fig:framework}(a) shows the channel: a sender in state $s$ writes $m$, the polisher $\Pi$ (drawn as a compressor) rewrites it, and a receiver $\rho$, a trained model or an LLM, reads both $m$ and $\Pi(m)$.

\subsection{Problem Statement}
\label{sec:problem}

A sender holds a communicative state $s\in\mathbb{R}$ along some social dimension, for instance how angry they are, and writes a message $m\sim P(m\mid s)$. A polisher $\Pi$ maps $m$ to $\Pi(m)$. A receiver $\rho$ observes a message and forms an estimate $\hat{s}=\rho(\cdot)$ of the sender's state. We take the annotated label of the original message as the operational $s$: it is the consensus of human readers of the original, and communication along this dimension succeeds to the extent that $\hat{s}$ tracks it. Prior measurements of homogenization characterize the marginal distribution of $\Pi(m)$ through its variance or lexical diversity, which has two limitations for our purpose. i) The variance of $\Pi(m)$ mixes the variation that encodes $s$ with variation that does not; a polisher that removes only the second kind helps the receiver and one that removes the first kind hurts it, and the marginal cannot tell them apart. ii) A receiver trained on $m$ and evaluated on $\Pi(m)$ conflates a change in the distribution of surface forms, which a reader exposed to polished text would learn to undo, with a many-to-one collapse of distinct states onto the same output, which no reader can undo. \oursname{} is built to separate the signal from the rest of the variation and the shift from the collapse.

\subsection{Measurements}
\label{sec:measures}

\paragraph{Retention.}
Editing intensity, against which we plot every loss, is the fraction of the original's tokens that do not survive in the polished text, computed per message and averaged per condition and dimension. For a receiver $\rho$ and a test set $\{(s_i,m_i)\}$, retention is the rank correlation between $\hat{s}$ and $s$ on polished text relative to the same correlation on original text,
\begin{equation}
\mathrm{Ret}(\Pi)=\mathrm{corr}\big(\rho(\Pi(m)),s\big)\,/\,\mathrm{corr}\big(\rho(m),s\big),
\label{eq:retention}
\end{equation}
with the area under the ROC curve in place of rank correlation for binary dimensions.

\paragraph{Compression Curve.}
Retention is blind to the shape of the loss. We therefore compare the receiver's reading of the original, $g(s)=\mathbb{E}[\rho(m)\mid s]$, with its reading of the polished version, $h(s)=\mathbb{E}[\rho(\Pi(m))\mid s]$, both estimated by binning $s$ into deciles and both expressed in units of the standard deviation of $s$. We summarize the departure the way an audio compressor is specified, by a knot $\tau$ and a slope on each side of it,
\begin{equation}
h\approx a+\beta_{-}\min(g-\tau,0)+\beta_{+}\max(g-\tau,0),
\label{eq:compressor}
\end{equation}
fitted by least squares with the knot searched over quantiles of $g$. Each dimension has a costly end, the top of the scale for toxicity, intensity, and distress and the bottom for politeness and empathy, and we report the slope on that side against the slope on the other, the shrinkage of the 5 to 95\% range of $\hat{s}$, and the displacement $h-g$ in the bottom and top quintiles of $s$, which exposes asymmetry.

\paragraph{Invertibility.}
We bound what a reader who adapts to polished text could recover by retraining the receiver on polished training data and evaluating it on polished test data. We treat the retention of this recalibrated receiver as an empirical ceiling, the most that a receiver with the same capacity and training budget recovers when it reads polished text as polished text; its gap to the original receiver on original text is the part of the signal that the channel has collapsed, and its gap to the original receiver on polished text is the part that was only shifted (Figure~\ref{fig:framework}(d): a shift moves every state by the same amount, so a recalibrated receiver recovers the states, while a collapse maps distinct states onto the same text). For LLM readers we use the same logic with an instruction: an informed reader is told that the message was polished by an LLM and asked to judge what the original writer felt.

\paragraph{Mechanism and Collective Consequences.}
For a set of surface features (profanity, intensifiers, exclamation marks, repeated punctuation, capitalization, elongation, hedges, emoji, and pronouns) we compute the removal rate, the fraction of messages in which a feature present in the original is absent after polishing, and we flag responses in which the polisher declines to rewrite. We then simulate a receiver who must rank a pool of senders by state, for instance to triage the most hostile, when a fraction $p$ of them polish: for each $p$ we draw the polishing senders at random, score every message with the receiver trained on original text, and report NDCG at the top decile against the true ranking and the recall of true top-decile senders by whether they polished, averaged over draws, together with the share of the scale, from the costly end, over which original messages are on average read as more extreme than polished messages are at any state.

\subsection{Instantiation}
\label{sec:setup}

\paragraph{Dimensions.}
Table~\ref{tab:data} lists the eight dimensions, drawn from seven datasets. Six are social signals with human annotation of the sender's state, graded for five of them and binary for irony: emotion intensity on tweets \citep{mohammad2018semeval}, politeness of requests \citep{danescu2013computational}, toxicity of comments \citep{borkan2019nuanced}, irony on tweets \citep{vanhee2018semeval}, and empathy and distress of reactions to news \citep{buechel2018modeling}, which share one set of messages. Two are content dimensions that a polisher should preserve and that serve as controls: sentiment polarity \citep{rosenthal2017semeval} and topic \citep{antypas2022twitter}. We subsample each training set to at most 4{,}000 messages and each test set to at most 1{,}500, stratified over the state so that the extremes are represented, and keep messages of 4 to 80 words (up to 200 for the empathy essays), 33{,}841 messages in all.

\paragraph{Polishers.}
We use four open-weight instruction-tuned models from two families: Qwen2.5-7B, Qwen3-8B, Qwen3-14B, and Llama-3.1-8B \citep{qwen2025qwen25,yang2025qwen3,grattafiori2024llama}. The main instruction is the one a user would type, ``Please polish this message before I send it. Keep its meaning.'' Three further instructions, run with Qwen3-8B, Qwen3-14B, and Llama-3.1-8B, probe realistic requests: grammar only, more professional, and a tone-preserving request that asks the model to keep the writer's exact tone, attitude, and emotional intensity. Decoding is greedy. As a control that paraphrases with no notion of good writing, we back-translate every message through German with Marian models \citep{tiedemann2020opus}.

\paragraph{Receivers.}
We fine-tune RoBERTa-large \citep{liu2019roberta} per dimension as the trained receiver, a regressor (or classifier for irony, sentiment, and topic) on the original training messages, with a held-out tenth for model selection; the recalibrated receiver uses the same recipe on polished training messages. The LLM reader is Qwen3-14B prompted zero-shot to rate a message on a 0 to 100 scale along the dimension (or to name its polarity or topic), either naively or with the informed instruction; it shares weights with one polisher, so the reader results in the body are under Llama-3.1-8B polishing. Full prompts and hyperparameters are in the appendix.

\begin{table}[t]
\centering
\footnotesize\setlength{\tabcolsep}{2.5pt}
\begin{tabular}{@{}llrrl@{}}
\toprule
Dimension & Source & Train & Test & State \\
\midrule
Emotion intensity & SemEval-18 T1 & 4{,}000 & 1{,}414 & $[0,1]$ \\
Politeness & Stanford & 4{,}000 & 1{,}500 & $z$-score \\
Toxicity & CivilComments & 4{,}000 & 1{,}500 & $[0,1]$ \\
Irony & SemEval-18 T3 & 3{,}786 & 781 & binary \\
Empathy/Distress & Buechel+18 & 1{,}488 & 372 & 1 to 7 \\
\midrule
Sentiment (control) & TweetEval & 4{,}000 & 1{,}500 & 3 classes \\
Topic (control) & TweetTopic & 4{,}000 & 1{,}500 & 6 classes \\
\bottomrule
\end{tabular}
\caption{Dimensions and data in \oursname{}. Test sets are the official development or test splits, subsampled with stratification over the state.}
\label{tab:data}
\end{table}

\section{Results}
\label{sec:results}

\subsection{Signal Retention}
\label{sec:retention}

Table~\ref{tab:retention} reports retention (Eq.~\ref{eq:retention}) for every dimension under the generic request to each polisher and under back-translation, measured with receivers trained on original text (Spearman correlation with the annotated state, AUC for irony and macro-F1 for the two classification controls; messages a polisher declined to rewrite are excluded for that polisher, and retention above 100\% means that the receiver ranks the polished messages slightly better than the originals). First, polishers differ in how much they rewrite, and the loss follows the rewriting. Qwen3-14B changes \neditQwenLGeneric\% of the tokens of a message under the generic request and retains \nretSocialLoQwenLGeneric{} to \nretSocialHiQwenLGeneric\% of the signal on the six social dimensions; Llama-3.1-8B changes \neditLlamaGeneric\% and retains \nretSocialLoLlamaGeneric{} to \nretSocialHiLlamaGeneric\%; Qwen2.5-7B and Qwen3-8B sit between them. Second, within a polisher the loss is selective. Under Llama-3.1-8B, toxicity retains \nrettoxicityLlamaGeneric\% and politeness \nretpolitenessLlamaGeneric\%, distress \nretdistressLlamaGeneric\% and emotion intensity \nretintensityLlamaGeneric\%, while irony and empathy retain \nretironyLlamaGeneric\% and \nretempathyLlamaGeneric\%, and the two content controls \nretsentimentLlamaGeneric\% and \nrettopicLlamaGeneric\%. Back-translation, which rewrites \neditBTBt\% of the tokens without any notion of good writing, retains \nretSocialLoBTBt{} to \nretSocialHiBTBt\% on the social dimensions, so paraphrase at a comparable editing intensity, with no model of good writing, does not produce the loss. Llama-3.1-8B under the generic request, whose editing intensity lies in the middle of the range we observe (\neditMin{} to \neditMax\% of tokens; Table~\ref{tab:edit} in the appendix), is the running example below, with Qwen3-14B as the mild case; Figure~\ref{fig:price} reports every condition, and the appendix compares the four requests.

\begin{table*}[t]
\centering
\footnotesize
\begin{tabular}{@{}lrrrrrr@{}}
\toprule
Dimension & Original & Qwen2.5-7B & Qwen3-8B & Qwen3-14B & Llama-3.1-8B & Back-transl. \\
\midrule
Emotion intensity & 0.771 & 0.732 (95\%) & 0.761 (99\%) & 0.759 (98\%) & 0.718 (93\%) & 0.717 (93\%) \\
Politeness & 0.575 & 0.516 (90\%) & 0.546 (95\%) & 0.558 (97\%) & 0.387 (67\%) & 0.510 (89\%) \\
Toxicity & 0.830 & 0.629 (76\%) & 0.747 (90\%) & 0.783 (94\%) & 0.345 (42\%) & 0.760 (92\%) \\
Irony & 0.805 & 0.808 (100\%) & 0.811 (101\%) & 0.816 (101\%) & 0.811 (101\%) & 0.780 (97\%) \\
Empathy & 0.452 & 0.489 (108\%) & 0.483 (107\%) & 0.477 (105\%) & 0.428 (95\%) & 0.453 (100\%) \\
Distress & 0.584 & 0.573 (98\%) & 0.571 (98\%) & 0.589 (101\%) & 0.503 (86\%) & 0.574 (98\%) \\
\midrule
Sentiment & 0.719 & 0.715 (99\%) & 0.715 (99\%) & 0.722 (100\%) & 0.716 (100\%) & 0.713 (99\%) \\
Topic & 0.876 & 0.869 (99\%) & 0.871 (99\%) & 0.866 (99\%) & 0.856 (98\%) & 0.855 (98\%) \\
\bottomrule
\end{tabular}

\caption{Retention of social signal and content under the generic request: correlation with the annotated state, with retention relative to original text in parentheses. Single seed; bootstrap intervals in Table~\ref{tab:bootstrap}.}
\label{tab:retention}
\end{table*}

\subsection{Cost of Editing}
\label{sec:price}

\begin{figure}[t]
\centering
\includegraphics[width=\columnwidth]{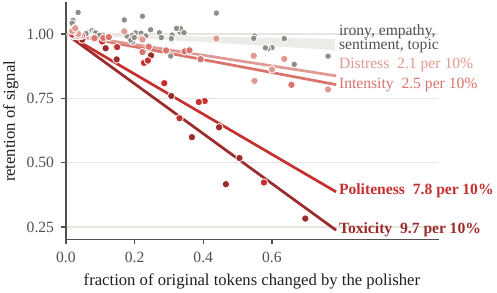}
\caption{Retention falls linearly with the fraction of original tokens a polisher changes, steeply for toxicity and politeness and barely for irony, empathy, sentiment, and topic. One point per polishing condition; receivers trained on original text, single seed.}
\label{fig:price}
\end{figure}

The two facts above combine into one measurement. For each of the fourteen conditions (four polishers under the generic request, three of them under the three further requests, and back-translation) we know how much of a message the polisher changed and how much of each signal survived; Figure~\ref{fig:price} plots one against the other (editing intensity is computed per dimension, so a condition sits at a different abscissa in each series; the lines are least-squares fits through full retention at zero edits, since an unedited message keeps its signal, and the four flattest signals are drawn as one grey band). Retention falls with editing intensity, but the exchange rate differs by an order of magnitude across dimensions: each ten percent of tokens changed costs \npriceTentoxicity{} points of toxicity retention (95\% bootstrap interval over conditions \npriceTenLotoxicity{} to \npriceTenHitoxicity{}, $R^2$ \npriceRsqtoxicity{}) and \npriceTenpoliteness{} points of politeness retention (\npriceTenLopoliteness{} to \npriceTenHipoliteness{}, $R^2$ \npriceRsqpoliteness{}), against \npriceTenintensity{} for emotion intensity, \npriceTendistress{} for distress, and under one point for irony, empathy, sentiment, and topic. Hostility and rudeness, the two dimensions on which a message can cost its sender something, are what the edits remove; content and the warmer signals pass through the same edits nearly intact.

Splitting the variance of the receiver's toxicity readings into a part explained by the annotated state and a remainder (Table~\ref{tab:variance} in the appendix), Llama-3.1-8B removes \nbetweenDroptoxicityLlamaGeneric\% of the signal and \nwithinDroptoxicityLlamaGeneric\% of the noise, the reverse of what an editor who removed only uninformative variation would do.

\subsection{Compression Curves}
\label{sec:compress}

\begin{figure*}[t]
\centering
\includegraphics[width=\textwidth]{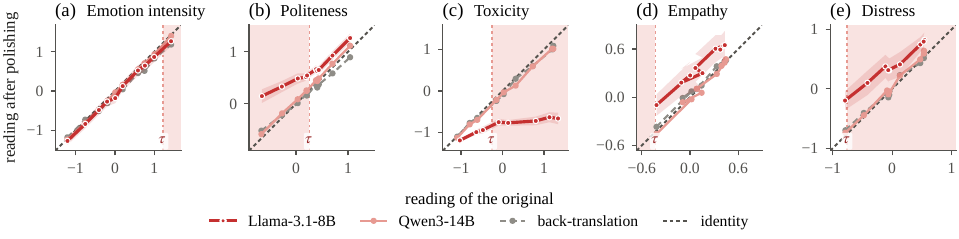}
\caption{Polishing flattens the readings of the most toxic and rudest messages and raises empathy and distress at every level. (a) to (e) Mean polished against mean original reading by state decile; single seed.}
\label{fig:compression}
\end{figure*}

Figure~\ref{fig:compression} shows the compression curves for the five graded dimensions (mean reading of the polished message against that of the original, by decile of the annotated state and in standard deviations of the state, with 95\% intervals for Llama-3.1-8B; Qwen3-14B in pink, back-translation in grey, and the costly side beyond the knot shaded; Figure~\ref{fig:asymmetry} in the appendix gives the displacement at each end). For toxicity the polished reading follows the original reading up to the knot and is flat beyond it: the slope of Eq.~\ref{eq:compressor} on the costly side is \nslopeCostlytoxicityLlamaGeneric{} against \nslopeSafetoxicityLlamaGeneric{} on the safe side, the 5 to 95\% range of the receiver's readings shrinks by \nrangetoxicityLlamaGeneric\% of its original width, and a message in the top quintile of annotated toxicity is read \nsoftenCostlytoxicityLlamaGeneric{} sd lower after polishing against \nsoftenSafetoxicityLlamaGeneric{} sd in the bottom quintile. Politeness shows the same shape from the other end: the rude quintile is read \nsoftenCostlypolitenessLlamaGeneric{} sd more polite and the polite quintile \nsoftenSafepolitenessLlamaGeneric{} sd, with slopes of \nslopeCostlypolitenessLlamaGeneric{} and \nslopeSafepolitenessLlamaGeneric{}. Emotion intensity stays close to the identity. Distress and empathy show a third pattern, a lift at every level: polished messages are read as more distressed by \nampSafedistressLlamaGeneric{} sd at the calm end and \nampCostlydistressLlamaGeneric{} sd at the distressed end, and as more empathetic by \nsoftenCostlyempathyLlamaGeneric{} sd at the low end and \nsoftenSafeempathyLlamaGeneric{} sd at the high end, which is consistent with the emotional vocabulary that LLM editing injects \citep{abdulhai2026how} seen from the receiver's side. Under Qwen3-14B every curve stays within \nmaxSoftenQwenLGeneric{} sd of the identity.

\subsection{Invertibility Analysis}
\label{sec:invert}

Table~\ref{tab:invert} in the appendix separates shift from collapse. For toxicity under Llama-3.1-8B a receiver retrained on polished text reaches a correlation of \nrecaltoxicityLlamaGeneric{} against \nshifttoxicityLlamaGeneric{} for the original receiver on polished text and \norigThreetoxicity{} on original text: it recovers \nrecovtoxicityLlamaGeneric\% of the gap, and its ceiling remains \nceilLosstoxicityLlamaGeneric\% below the original. For politeness the retrained receiver recovers \nrecovpolitenessLlamaGeneric\% of the gap and stays \nceilLosspolitenessLlamaGeneric\% below; for emotion intensity and distress retraining recovers nothing (Figure~\ref{fig:pairwise} in the appendix shows the same for pairs of messages). The zero-shot LLM reader, which shares no training with the receivers, shows the same selectivity (Table~\ref{tab:reader} in the appendix): under Llama-3.1-8B its correlation with annotated toxicity falls from \nreaderOrigtoxicity{} to \nreaderNaivetoxicityLlamaGeneric{} and with politeness from \nreaderOrigpoliteness{} to \nreaderNaivepolitenessLlamaGeneric{}, while emotion intensity keeps \nreaderRetintensityLlamaGeneric\%. Telling it that the message was polished, and asking it to judge the original writer, moves its agreement by at most \nreaderInformedDeltaMaxLlamaGeneric{} points on any dimension and closes \nreaderGapClosedtoxicityLlamaGeneric\% of the toxicity gap, against \nreaderGapClosedtoxicityBTBt\% of the smaller gap that back-translation opens. Part of the loss is a shift that an adapted reader can learn to undo; the larger part, for the two costly dimensions, is a collapse of distinct states onto the same text, which no knowledge of the channel undoes.

\begin{figure*}[t]
\centering
\includegraphics[width=\textwidth]{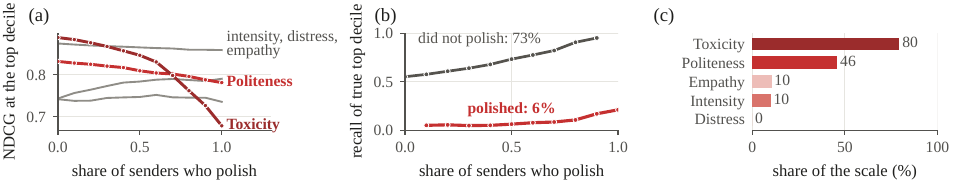}
\caption{As more senders polish, rankings by toxicity and politeness degrade and miss those who polish. (a) NDCG at the top decile. (b) Recall of hostile senders by whether they polished. (c) Scale beyond polished readings.}
\label{fig:collective}
\end{figure*}

\subsection{Mechanism Analysis}
\label{sec:mechanism}

\begin{figure}[t]
\centering
\includegraphics[width=\columnwidth]{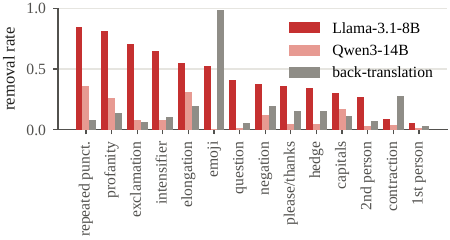}
\caption{Polishers remove the surface forms that carry social cost. Removal rate of each feature: the fraction of test messages containing it in which it is absent after rewriting, averaged over the six social dimensions and ordered by the rate under Llama-3.1-8B.}
\label{fig:mechanism}
\end{figure}

Figure~\ref{fig:mechanism} shows, for the surface features that style guides treat as excess, how often a polisher deletes them. Llama-3.1-8B removes repeated punctuation from \nremovalmultipunctLlamaGeneric\% of the messages that contain it, profanity from \nremovalprofanityLlamaGeneric\%, exclamation marks from \nremovalexclamationLlamaGeneric\%, intensifiers from \nremovalintensifierLlamaGeneric\%, and elongated words from \nremovalelongationLlamaGeneric\%, while removing first-person pronouns from only \nremovalfirstpersonLlamaGeneric\% of theirs; Qwen3-14B removes profanity from \nremovalprofanityQwenLGeneric\% and exclamation marks from \nremovalexclamationQwenLGeneric\%. Profanity alone has a rank correlation of \nrhoProfanityTox{} with annotated toxicity, which is why its removal costs so much of that signal. The polisher is doing what it was asked to do: in human writing the carriers of social cost are, by convention, exactly the forms a style guide calls excessive. A polisher cannot tell a message that is written carelessly from one that is written hostilely, because on the surface they are the same message. The same instruction has a second effect that is easy to miss: Llama-3.1-8B declines to rewrite \nrefusetoxicityLlamaGeneric\% of the toxicity messages and \nrefusesentimentLlamaGeneric\% of the sentiment tweets, almost all at the hostile or explicit end of the scale (Table~\ref{tab:refusals} in the appendix), so that the most extreme senders are silenced instead of softened; the Qwen models decline at most 0.1\% of any dimension.

\subsection{Collective Consequences}
\label{sec:collective}

Figure~\ref{fig:collective} simulates a receiver who must rank a pool of senders by state when a fraction $p$ of them polish with Llama-3.1-8B (100 random draws per point; toxicity and politeness in red, the other dimensions in grey). NDCG at the top decile of toxicity falls from \nndcgZerotoxicityLlamaGeneric{} at $p=0$ to \nndcgHalftoxicityLlamaGeneric{} at $p=0.5$ and \nndcgOnetoxicityLlamaGeneric{} at $p=1$, and for politeness from \nndcgZeropolitenessLlamaGeneric{} to \nndcgOnepolitenessLlamaGeneric{}; emotion intensity and distress barely move, and empathy, which the polisher amplifies, becomes easier to rank. Figure~\ref{fig:collective}(b) shows that the loss is not spread evenly. At $p=0.5$, a sender in the true top decile of toxicity is recovered \nrecallUnpoltoxicityLlamaGeneric\% of the time if they did not polish and \nrecallPoltoxicityLlamaGeneric\% of the time if they did: polishing moves a sender down the receiver's ranking, so the receiver's errors fall on the senders who adopt the tool. Figure~\ref{fig:collective}(c) inverts the compression curve: the mean reading of a polished message at the most toxic state lies below the mean reading of original messages over the top \nunreachtoxicity\% of the toxicity scale, and over the rudest \nunreachpoliteness\% of politeness.

\section{Discussion}
\label{sec:discussion}

\paragraph{Compressor Specification.}
Read as a compressor, the loss is located in units comparable across dimensions, and the reading reconciles prior lexical measurements with ours: a compressor raises the average level of a signal while reducing its excursion, which is what an LLM that adds emotional vocabulary to every message \citep{abdulhai2026how} while erasing the difference between a calm and a furious one does.

\paragraph{Tone-Preserving Requests.}
A request to preserve tone (Table~\ref{tab:prompts} in the appendix) is obeyed by one model and ignored by another, and where it is obeyed it works by changing almost nothing. This is what the mechanism in Section~\ref{sec:mechanism} predicts: the variation that carries social cost and the variation that carries nothing share the same surface forms, and a model cannot keep the first and discard the second without a model of the sender, which neither the instruction nor the text provides. Social dynamic range is therefore a property to measure and train for, as propositional content is.

\paragraph{Downstream Consequences.}
Any platform that triages messages by hostility will see the ranking loss of Section~\ref{sec:collective}; corpus studies inherit text whose variance along these dimensions has been compressed in proportion to adoption; and individual readers lose the signals they use to decide whom to help and how urgently, without any indication that the signal was ever there. Every collective quantity in Section~\ref{sec:collective} scales with the share of messages that pass through a polisher, and that share is still rising \citep{liang2025widespread,sun2025are}.

\section{Conclusion}

We have asked what a reader can still learn about a sender from a message an LLM has polished, and answered with a channel model and an evaluation suite, \oursname{}, that measures the social signal surviving the channel. Polishing compresses the dynamic range of hostility and rudeness at the costly end of each scale while irony, empathy, topic, and sentiment pass through; a reader who adapts undoes only part of the loss, because the forms that carry the signal are the ones a polisher is asked to remove; and as more senders polish, the senders a receiver misses are the ones who polish. We hope these early measurements make preserving social signal an explicit objective for writing assistants and a routine check for anyone who reads populations through text.

\section*{Limitations}

Although \oursname{} measures the social signal that survives polishing across six dimensions, four polishers, and four requests, its receivers are models: the trained receiver stands for the automated triage already run on text of this kind, and the LLM reader for a capable reader with no exposure to polished text. The ground truth is therefore a human annotation of the original message, and a study with human readers of polished text would complete the picture. The dimensions and polishers are English and open-weight, chosen so that every rewrite is reproducible, and each message is polished in isolation with greedy decoding, so proprietary polishers, other languages, and the conversational context of a real request may compress differently; we have no evidence that they compress less. Because \oursname{} is agnostic to the polisher, the exchange rate of Section~\ref{sec:price} is the quantity on which these settings can be compared next.

\section*{Broader Impact and Ethical Considerations}

Writing assistants are deployed as editors of form, and our results show that they also decide how much of a sender's state reaches the reader. The same view applies to any rewriting system placed between people, such as machine translation, message summarization, and suggested replies, and it implies that such systems should be evaluated on the receiver's inference about the sender and trained with an explicit objective of preserving social dynamic range, for instance by penalizing the collapse that a recalibrated receiver cannot undo. On the negative side, the receivers and measurements in \oursname{} could be used to tune a polisher that hides hostility from moderators, and recovering a sender's emotional state from text can be used against the sender, for whom polishing offers a measure of privacy; conversely, the same compression removes the signals that let a reader recognize distress or urgency. We therefore report both effects so that platforms and users can weigh them, and we release only the measurement suite. All data are public research datasets used under their original licenses; we add no new annotation, release no raw text beyond the sources, involve no human subjects, and process offensive comments only to measure what a polisher removes.

\bibliography{refs}

\appendix

\section{Protocol Overview}
\label{app:overview}
\begin{figure*}[t]
\centering
\includegraphics[width=\textwidth]{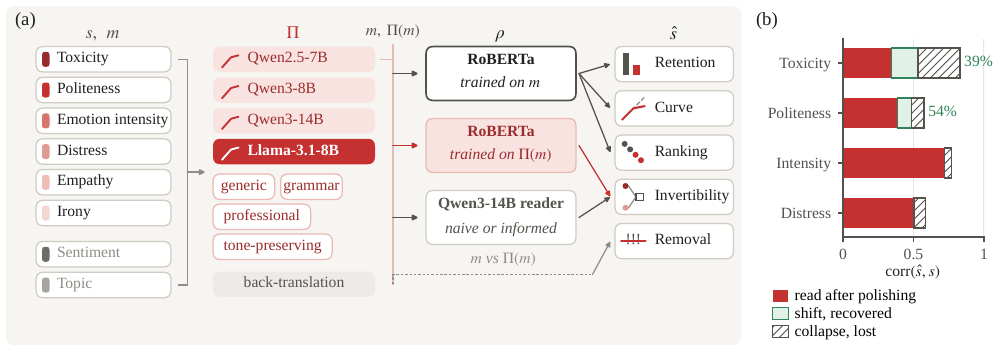}
\caption{\oursname{} separates the shift that a reader can undo from the collapse that it cannot. (a) Data, channels, receivers, and what each measurement compares. (b) Shift and collapse in the receiver's loss under Llama-3.1-8B.}
\label{fig:protocol}
\end{figure*}

Figure~\ref{fig:protocol}(a) traces a test message through \oursname{}. Each of the eight dimensions (the two controls in grey) is rewritten by four polishers, under the generic request for all four and under the grammar, professional, and tone-preserving requests for Qwen3-8B, Qwen3-14B, and Llama-3.1-8B, and by back-translation; three receivers read the original and the polished text, and each measurement uses the comparison drawn into it, with the removal rates read off the two texts without a receiver. Figure~\ref{fig:protocol}(b) applies the invertibility measurement to Llama-3.1-8B: the red part of each bar is the correlation of the original receiver on polished text, the green part is the gain of the receiver retrained on polished text, which is the shift, and the hatched part is the remaining distance to the original correlation, which is the collapse (the percentages give the recovered share of the gap; for emotion intensity and distress retraining recovers nothing).

\section{Prompts}
\label{app:prompts}
The polishing requests are sent as the user turn of a chat with the system prompt ``You are a helpful writing assistant.'' and the message appended after ``Message:''. The four requests are: \emph{generic}, ``Please polish this message before I send it. Keep its meaning. Output only the polished message, nothing else.''; \emph{grammar}, ``Fix only the grammar, spelling and punctuation of this message. Do not change anything else. Output only the corrected message, nothing else.''; \emph{professional}, ``Rewrite this message so it sounds more professional. Output only the rewritten message, nothing else.''; and \emph{tone-preserving}, ``Polish this message, but preserve the writer's exact tone, attitude and emotional intensity, including how angry, polite, sarcastic or distressed it sounds. Output only the polished message, nothing else.'' Thinking mode is disabled for Qwen3 models. Outputs are stripped of surrounding quotation marks and of any preamble such as ``Here is the polished message:''.

The LLM reader receives the system prompt ``You are a careful reader of social media messages. Answer only with what is asked, no explanation.'' and a question per dimension, for instance ``How intense is the anger felt by the writer of this message? Answer with a single integer from 0 (no anger at all) to 100 (the most intense anger imaginable).'' The informed variant prepends ``Note: before being sent, this message was polished by an AI writing assistant (a large language model), which may have changed its wording and tone. Judge what the ORIGINAL writer felt and intended, not how the polished text sounds.''

\section{Receiver Training}
\label{app:receivers}
RoBERTa-large is fine-tuned for five epochs with AdamW, learning rate $10^{-5}$, batch size 16, weight decay 0.01, linear warm-up over the first 10\% of steps, gradient clipping at 1.0, and bfloat16 autocast, with the epoch of best held-out Spearman correlation (macro-F1 for classification) retained. Regression targets are standardized on the training set. The input for emotion intensity is prefixed with the emotion whose intensity is annotated. Maximum sequence length is 128 tokens, or 256 for the empathy essays. Every receiver is trained once, with a single random seed; the uncertainty we report comes from the test messages. Table~\ref{tab:bootstrap} gives 95\% intervals for retention from 1{,}000 paired bootstrap resamples of the test set, in which the original and the polished reading of a message are drawn together and messages the polisher refused are excluded from both. The bootstrap standard error of retention is at most \nbootSeMaxLarge{} points on the six dimensions with 700 or more test messages and at most \nbootSeMaxSmall{} points on empathy and distress, whose test set has 372 essays; the paired estimate differs from the ratio of the two unpaired correlations reported in Table~\ref{tab:retention} by at most \nbootPairedDiffMax{} points.

\section{Additional Results}
\label{app:more}
\begin{figure*}[t]
\centering
\includegraphics[width=\textwidth]{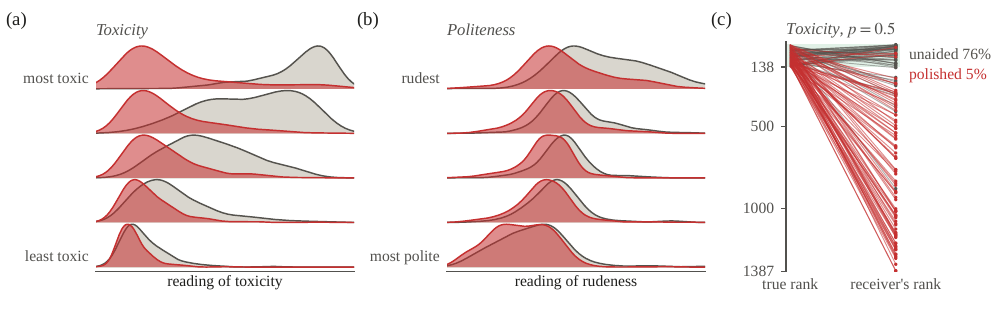}
\caption{Polishing merges the readings of different states, and the errors fall on the senders who polish. (a, b) Readings by quintile of the annotated state. (c) True and estimated rank of the most toxic senders in one draw.}
\label{fig:accounting}
\end{figure*}

\begin{figure*}[t]
\centering
\includegraphics[width=\textwidth]{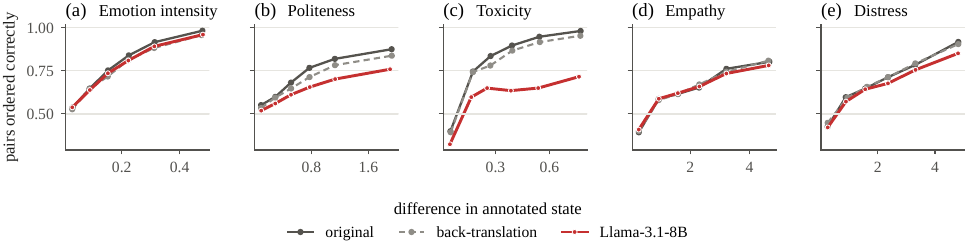}
\caption{Polishing makes pairs harder to order on toxicity and politeness and leaves the other signals nearly unchanged. (a) to (e) Fraction of random test pairs the receiver orders correctly, by difference in annotated state.}
\label{fig:pairwise}
\end{figure*}

\begin{figure}[h]
\centering
\includegraphics[width=\columnwidth]{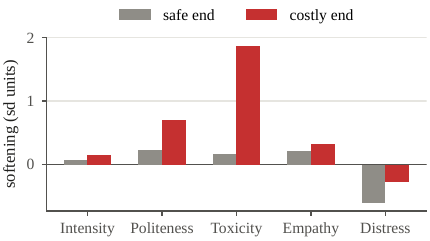}
\caption{Polishing displaces the reading more at the costly end of toxicity and politeness than at the safe end. Mean displacement of the polished reading from the original reading in the bottom and top quintile of each dimension, in sd of the state (Llama-3.1-8B).}
\label{fig:asymmetry}
\end{figure}

Table~\ref{tab:refusals} gives the refusals, detected with a pattern over the response; back-translation produces none and the Qwen models at most 0.1\% of any dimension, and treating the refusal text as the delivered message instead of excluding it lowers the toxicity retention of Llama-3.1-8B from \nrettoxicityLlamaGeneric\% to \nretInclRefusetoxicityLlamaGeneric\%. In Table~\ref{tab:invert}, Recov. is the fraction of the gap between the shifted and the original correlation that the recalibrated receiver closes, shown when the gap exceeds 0.02. Table~\ref{tab:variance} splits the variance of the readings with quintile bins of the annotated state, on original text, after Llama-3.1-8B polishing, and after back-translation. Table~\ref{tab:reader} gives the zero-shot reader, whose informed variant is told that the message was polished and asked to judge the original writer. Figure~\ref{fig:pairwise} compares original text (grey), Llama-3.1-8B (red), and back-translation (grey dashed). Table~\ref{tab:edit} gives the editing intensity of every condition, and Figure~\ref{fig:asymmetry} the displacement of the polished reading at each end of the scale.

Figure~\ref{fig:accounting} shows the same loss at the level of single readings and of a ranked pool. Figure~\ref{fig:accounting}(a) and (b) give kernel densities of the original receiver's readings in each quintile of the annotated state, on original text (grey) and after Llama-3.1-8B polishing (red), with the costly quintile on top. On original text the quintiles of toxicity separate; after polishing the three most toxic quintiles fall on nearly the same distribution, and for politeness the rudest quintile is moved onto its neighbours. Figure~\ref{fig:accounting}(c) takes one pool from the ranking simulation of Section~\ref{sec:collective} at $p=0.5$ and links the true and estimated rank of every sender in the true top decile of toxicity: in this draw 76\% of unaided and 5\% of polished senders stay in the top decile, against averages of \nrecallUnpoltoxicityLlamaGeneric\% and \nrecallPoltoxicityLlamaGeneric\% over 100 draws.

\begin{table}[h]
\centering
\footnotesize\setlength{\tabcolsep}{2.5pt}
\begin{tabular}{@{}llrrrr@{}}
\toprule
Dimension & Polisher & Orig. & Shifted & Recal. & Recov. \\
\midrule
Intensity & Llama-3.1-8B & 0.771 & 0.718 & 0.718 & 0.00 \\
 & Back-transl. & 0.771 & 0.717 & 0.680 & 0.00 \\
Politeness & Llama-3.1-8B & 0.575 & 0.387 & 0.488 & 0.54 \\
 & Back-transl. & 0.575 & 0.510 & 0.527 & 0.25 \\
Toxicity & Llama-3.1-8B & 0.830 & 0.345 & 0.533 & 0.39 \\
 & Back-transl. & 0.830 & 0.760 & 0.763 & 0.04 \\
Irony & Llama-3.1-8B & 0.805 & 0.811 & 0.852 & -- \\
 & Back-transl. & 0.805 & 0.780 & 0.872 & 1.00 \\
Empathy & Llama-3.1-8B & 0.452 & 0.428 & 0.430 & 0.07 \\
 & Back-transl. & 0.452 & 0.453 & 0.483 & -- \\
Distress & Llama-3.1-8B & 0.584 & 0.503 & 0.496 & 0.00 \\
 & Back-transl. & 0.584 & 0.574 & 0.557 & -- \\
\bottomrule
\end{tabular}

\caption{Invertibility: original receiver on original (Orig.) and polished text (Shifted), receiver retrained on polished text (Recal.). Spearman, or AUC for irony.}
\label{tab:invert}
\end{table}

\begin{table*}[h]
\centering
\small
\begin{tabular}{@{}lccccc@{}}
\toprule
Dimension & Qwen3-14B & Qwen3-8B & Qwen2.5-7B & Llama-3.1-8B & Back-transl. \\
\midrule
Emotion intensity & 98 [97, 99] & 99 [98, 100] & 95 [93, 97] & 94 [91, 96] & 93 [91, 95] \\
Politeness & 97 [95, 99] & 95 [92, 97] & 90 [85, 93] & 67 [61, 74] & 89 [85, 93] \\
Toxicity & 94 [92, 96] & 90 [87, 92] & 76 [72, 80] & 42 [36, 47] & 92 [89, 94] \\
Empathy & 105 [99, 114] & 107 [100, 115] & 108 [99, 120] & 97 [86, 110] & 100 [93, 109] \\
Distress & 101 [96, 106] & 98 [92, 103] & 98 [92, 105] & 87 [77, 98] & 98 [93, 103] \\
Irony & 101 [99, 103] & 101 [99, 103] & 100 [98, 103] & 101 [98, 104] & 97 [94, 100] \\
\midrule
Sentiment (control) & 100 [98, 102] & 99 [98, 101] & 99 [97, 102] & 99 [96, 102] & 99 [97, 101] \\
Topic (control) & 99 [98, 100] & 99 [99, 100] & 99 [98, 100] & 98 [96, 99] & 98 [96, 99] \\
\bottomrule
\end{tabular}

\caption{Retention (\%) with 95\% intervals from 1{,}000 paired bootstrap resamples of the test messages; single seed.}
\label{tab:bootstrap}
\end{table*}

\paragraph{Requests and Models.}
Table~\ref{tab:prompts} varies the request. A grammar-only request changes \neditQwenLGrammar{} to \neditLlamaGrammar\% of the tokens and retains nearly all of the signal; the professional request is the most destructive, with Llama-3.1-8B changing \neditLlamaProfessional\% of the tokens and retaining \nrettoxicityLlamaProfessional\% of toxicity and \nretpolitenessLlamaProfessional\% of politeness. The tone-preserving request, which names the very quantities we measure, is obeyed by Qwen3-14B, which then changes \neditQwenLPreserve\% of the tokens, and largely ignored by Llama-3.1-8B, which still changes \neditLlamaPreserve\% but retains \nrettoxicityLlamaPreserve\% of the toxicity signal against \nrettoxicityLlamaGeneric\% under the generic request at the same intensity. The request mostly moves a polisher along Figure~\ref{fig:price}; the exchange rate is set by the polisher's edits and modulated by the request.

\begin{table*}[h]
\centering
\small
\begin{tabular}{@{}llrrrrrrr@{}}
\toprule
Model & Request & Edit & Intens. & Polite. & Toxic. & Irony & Empathy & Distress \\
\midrule
Qwen3-14B & grammar & 3\% & 99 & 99 & 100 & 100 & 105 & 101 \\
 & generic & 10\% & 98 & 97 & 94 & 101 & 105 & 101 \\
 & tone-preserving & 2\% & 99 & 99 & 100 & 101 & 104 & 101 \\
 & professional & 29\% & 94 & 81 & 60 & 102 & 98 & 91 \\
\midrule
Llama-3.1-8B & grammar & 6\% & 98 & 99 & 98 & 101 & 108 & 100 \\
 & generic & 36\% & 93 & 67 & 42 & 101 & 95 & 86 \\
 & tone-preserving & 37\% & 94 & 74 & 64 & 102 & 99 & 82 \\
 & professional & 63\% & 80 & 42 & 28 & 88 & 91 & 78 \\
\bottomrule
\end{tabular}

\caption{Retention (\%) under four requests for Qwen3-14B and Llama-3.1-8B, with the fraction of tokens changed.}
\label{tab:prompts}
\end{table*}

\begin{table}[t]
\centering
\footnotesize\setlength{\tabcolsep}{2.5pt}
\begin{tabular}{@{}lrr@{}}
\toprule
Polisher / request & Changed & Len. ratio \\
\midrule
Qwen3-14B / generic & 10\% & 1.00 \\
Qwen3-14B / grammar & 3\% & 1.00 \\
Qwen3-14B / professional & 29\% & 1.05 \\
Qwen3-14B / tone-preserving & 2\% & 1.00 \\
Qwen3-8B / generic & 13\% & 1.00 \\
Qwen3-8B / grammar & 2\% & 1.00 \\
Qwen3-8B / professional & 39\% & 1.06 \\
Qwen3-8B / tone-preserving & 2\% & 1.00 \\
Qwen2.5-7B / generic & 25\% & 1.00 \\
Llama-3.1-8B / generic & 36\% & 1.12 \\
Llama-3.1-8B / grammar & 6\% & 1.03 \\
Llama-3.1-8B / professional & 63\% & 1.51 \\
Llama-3.1-8B / tone-preserving & 37\% & 1.35 \\
back-translation & 21\% & 0.98 \\
\bottomrule
\end{tabular}

\caption{Editing intensity: mean fraction of original tokens changed and median length ratio of polished to original messages, per condition.}
\label{tab:edit}
\end{table}

\begin{table}[h]
\centering
\footnotesize\setlength{\tabcolsep}{3pt}
\begin{tabular}{@{}lrrrrrr@{}}
\toprule
& \multicolumn{3}{c}{Between (signal)} & \multicolumn{3}{c}{Within (noise)} \\
Dimension & Orig. & Llama & BT & Orig. & Llama & BT \\
\midrule
Emotion intensity & 0.58 & 0.53 & 0.46 & 0.44 & 0.55 & 0.48 \\
Politeness & 0.19 & 0.08 & 0.13 & 0.39 & 0.33 & 0.37 \\
Toxicity & 0.66 & 0.04 & 0.58 & 0.33 & 0.29 & 0.44 \\
Empathy & 0.09 & 0.07 & 0.09 & 0.33 & 0.28 & 0.31 \\
Distress & 0.21 & 0.11 & 0.19 & 0.42 & 0.33 & 0.39 \\
\bottomrule
\end{tabular}

\caption{Variance of the receiver's readings (squared sd of the state), split into the part explained by the annotated state and the remainder.}
\label{tab:variance}
\end{table}

\begin{table}[h]
\centering
\footnotesize\setlength{\tabcolsep}{2.5pt}
\begin{tabular}{@{}llrrr@{}}
\toprule
Dimension & Polisher & Original & Naive & Informed \\
\midrule
Intensity & Qwen3-14B & 0.550 & 0.550 & 0.562 \\
 & Llama-3.1-8B & 0.550 & 0.538 & 0.565 \\
 & Back-transl. & 0.550 & 0.525 & 0.541 \\
Politeness & Qwen3-14B & 0.385 & 0.414 & 0.412 \\
 & Llama-3.1-8B & 0.385 & 0.241 & 0.237 \\
 & Back-transl. & 0.385 & 0.321 & 0.315 \\
Toxicity & Qwen3-14B & 0.523 & 0.497 & 0.524 \\
 & Llama-3.1-8B & 0.523 & 0.190 & 0.208 \\
 & Back-transl. & 0.523 & 0.489 & 0.515 \\
Irony & Qwen3-14B & 0.866 & 0.865 & 0.886 \\
 & Llama-3.1-8B & 0.866 & 0.802 & 0.811 \\
 & Back-transl. & 0.866 & 0.783 & 0.808 \\
Empathy & Qwen3-14B & 0.375 & 0.380 & 0.405 \\
 & Llama-3.1-8B & 0.375 & 0.292 & 0.319 \\
 & Back-transl. & 0.375 & 0.367 & 0.386 \\
Distress & Qwen3-14B & 0.455 & 0.467 & 0.469 \\
 & Llama-3.1-8B & 0.455 & 0.430 & 0.390 \\
 & Back-transl. & 0.455 & 0.459 & 0.444 \\
\bottomrule
\end{tabular}

\caption{Zero-shot LLM reader (Qwen3-14B) on original and polished text, naive or informed. Spearman correlation with the annotated state, or AUC for irony.}
\label{tab:reader}
\end{table}

\begin{table}[h]
\centering
\footnotesize\setlength{\tabcolsep}{3pt}
\begin{tabular}{@{}lrrrr@{}}
\toprule
& \multicolumn{3}{c}{Llama-3.1-8B} & Qwen3-14B \\
Dimension & generic & profess. & tone & generic \\
\midrule
Emotion intensity & 2.1 & 1.3 & 0.6 & 0.0 \\
Politeness & 0.1 & 0.1 & 0.1 & 0.0 \\
Toxicity & 7.5 & 5.3 & 1.1 & 0.0 \\
Irony & 2.2 & 1.9 & 0.4 & 0.0 \\
Empathy & 1.6 & 1.6 & 0.5 & 0.0 \\
Sentiment & 4.4 & 4.5 & 0.5 & 0.0 \\
Topic & 0.6 & 0.6 & 0.0 & 0.0 \\
\bottomrule
\end{tabular}

\caption{Share (\%) of test messages that the polisher declined to rewrite, by request (generic, professional, tone-preserving) and dimension.}
\label{tab:refusals}
\end{table}

\end{document}